%% file: main.tex
\documentclass[conference]{IEEEtran}
\IEEEoverridecommandlockouts
\usepackage[T1]{fontenc}
\usepackage[latin1]{inputenc}
\usepackage{subfig}
\usepackage{float}
\usepackage{graphicx}
\usepackage{setspace}
\usepackage{cite}
\usepackage{babel}
\usepackage{amsmath,amssymb,amsfonts}
\usepackage{acronym}
\usepackage{url}

\usepackage{xcolor}
\usepackage{listings}
\usepackage[separate-uncertainty=true]{siunitx}
\title{SNNAX - Spiking Neural Networks in JAX}

\author{\IEEEauthorblockN{Jamie Lohoff$^*$}
\IEEEcompsocitemizethanks{* These authors contributed equally.}
\IEEEauthorblockA{\textit{Peter-Gr\"unberg Institute 15} \\
\textit{FZ J\"ulich, RWTH Aachen} \\
Aachen, Germany \\
ja.lohoff@fz-juelich.de}
\and
\IEEEauthorblockN{Jan Finkbeiner$^*$}
\IEEEauthorblockA{\textit{Peter-Gr\"unberg Institute 15} \\
\textit{FZ J\"ulich, RWTH Aachen}\\
Aachen, Germany \\
j.finkbeiner@fz-juelich.de}
\and
\IEEEauthorblockN{Emre Neftci$^*$}
\IEEEauthorblockA{\textit{Peter-Gr\"unberg Institute 15} \\
\textit{FZ J\"ulich, RWTH Aachen} \\
Aachen, Germany \\
e.neftci@fz-juelich.de}}

\begin{document}

\input{acronyms}

\maketitle

\begin{abstract}
    \acp{SNN} simulators are essential tools to prototype biologically inspired models and neuromorphic hardware architectures and predict their performance.
    For such a tool, ease of use and flexibility are critical, but so is simulation speed especially given the complexity inherent to simulating \ac{SNN}.
    Here, we present SNNAX, a JAX-based framework for simulating and training such models with PyTorch-like intuitiveness and JAX-like execution speed. 
    SNNAX models are easily extended and customized to fit the desired model specifications and target neuromorphic hardware.
    Additionally, SNNAX offers key features for optimizing the training and deployment of \acp{SNN} such as flexible automatic differentiation and just-in-time compilation.
    We evaluate and compare SNNAX to other commonly used \ac{ML} frameworks used for programming SNNs.
    We provide key performance metrics, best practices, documented examples for simulating \acp{SNN} in SNNAX, and implement several benchmarks used in the literature.
\end{abstract}

\bstctlcite{BSTcontrol} % DO NOT DELETE

\section{Introduction}
\acs{SNN} simulators leveraging parallel processing have become indispensable tools to rapidly evaluate and test their performance.
Furthermore, as most neuromorphic hardware are based on spiking neurons, such simulators have become essential to evaluate the performance of prototypes at large scale.
Many state-of-the-art \acp{SNN} are commonly trained with gradient-based learning; typically facilitated through a modern machine learning (\ac{ML}) framework like PyTorch or JAX \cite{Bradbury_etal18_jaxcomp, Paszke_etal17_autodiff} and parallel computing accelerators like GPUs and TPUs.
These frameworks are highly modular and extensible and thus a popular choice for a dynamic and growing field such as neuromorphic computing.
Furthermore, \ac{ML} frameworks are optimized for performance, allowing for efficient training of potentially large-scale SNN models on modern hardware accelerators.
A host of recent work leveraged \ac{ML} frameworks for simulating \acp{SNN} \cite{Eshraghian_etal23_traispik,Pehle_Pedersen21_nors, wei2023_spikj, heckel2024_spyx, muir2019_rockpool, sheik2021_sinabs}.

Here, we report SNNAX, our JAX-based library for simulating \acp{SNN} that is built on Equinox \cite{kidger2021_equinox}, a thin neural network and numerical computation library.
We chose JAX as the underlying \ac{ML} framework because it includes several features that are essential for algorithmic exploration while offering high execution performance. 
Exploring novel learning algorithms is essential to address challenges arising from backpropagation-based methods, \emph{e.g.} by incorporating results from synaptic plasticity \cite{Zenke_Neftci21_brailear}. 
This type of algorithmic exploration is enabled by JAX's extensive automatic differentiation library, its Just-In-Time (JIT) compiler based on XLA, and a paradigm that separates functions and parameters (functional programming).
We compare the effectiveness of SNNAX and similar SNN libraries provide notebooks implementing the standard benchmarks inspired by the Neurobench initiative \cite{yik2024neurobench}.

\section{Related Work}
Due to neuromorphic computing borrowing from both Neuroscience and AI, there is a significant functional overlap between \ac{SNN} "brain" simulators like NEST\cite{Gewaltig2007_NEST}, Brian2 \cite{Stimberg2019_brian2} and GeNN \cite{yavuz2016genn} and SNN libraries for machine learning frameworks.
While these brain simulators enable distributed, highly scalable simulations on supercomputers, their main design principles are biological accuracy and reproducibility and are thus currently unsuited to explore novel AI applications.
Missing in such frameworks are traceable dynamic variables that enables the computation of gradients and thus the efficient training of \acp{SNN}.\\
%The main reason being the still unchallenged superiority of gradient-based learning over any other approach even for more bio-plausible SNNs which has only limited support within these libraries.
Thus, most SNN libraries are built around an automatic differentiation (AD) library like PyTorch or JAX that allows for easy and efficient gradient computations.
Popular library choices for PyTorch include \textit{snnTorch}, \textit{Norse} and \textit{Spiking Jelly} \cite{Eshraghian_etal23_traispik, Pehle_Pedersen21_nors, fang2023spikjel}.
\textit{snnTorch} leverages PyTorch's own compiler to accelerate the otherwise impractically slow training of large-scale \acp{SNN} in Python.
\textit{Norse} also builds on PyTorch but instead of discretizing incoming spikes into a fixed temporal grid, it leverages an event-driven simulation scheme.
Both capitalize on many of the already implemented ML features like optimizers, data loading and popular neural network layers such as convolutions.\\
\textit{Spiking Jelly} leverages custom CUDA kernels for the spiking neuron dynamics implemented through the \textit{CuPy} interface to further accelerate the ML workloads.
It supports both layer-by-layer and step-by-step execution of the neuron models, thereby giving a lot of flexibility to the user.
\textit{SINABS} \cite{sheik2021_sinabs} is another example for a PyTorch-based library that is functionally similar to snnTorch, but with a focus on neuromorphic hardware.
Gradient computations can be further accelerated with the EXODUS extension \cite{bauer2022exodus} which uses implicit differentiation and custom CUDA kernels for the SRM model.\newpage
\begin{figure}
\centering
  \includegraphics[width=0.44\textwidth]{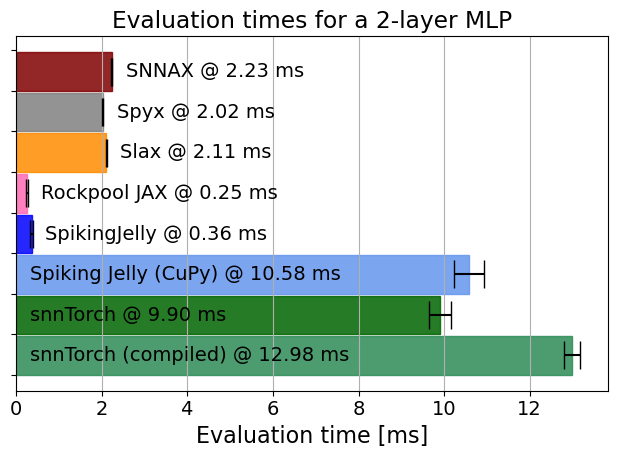}
\caption{Execution time measurements of a two-layer MLP with blocks of 2048 LIF neurons for batchsize 32.}
\label{fig:mlp}
\end{figure}
However, these features only support also only feed-forward architectures and a limited class of neurons.
%Two of it's main advantages are that it comes with two supporting SNN inference platforms, the DynapCNN chip and the Speck chip and also provides a CUDA implementation of the Spike-Response-Model (SRM) and it's 
A similar neuromorphic-hardware focused library is \textit{SLAYER} \cite{Shrestha_Orchard18_slayspik}, which includes a functional simulation of the Intel Loihi and custom CUDA kernels for the computation of delay gradients.
% However, with the introduction of compilation for PyTorch computational graphs, the performance gap to JAX narrowed down significantly.
JAX is a popular alternative to PyTorch which leverages a functional programming paradigm. 
Treating many of its features as function transformations, JAX puts emphasis on composability and uses the model's source code as a template to generate new source code that supports JIT compilation, vectorization or auto-differentiability.
JAX itself does not provide convenience functions and neural network building blocks, but there exist several neural network libraries with different design paradigms such as Haiku, Flax and Equinox.
% Massive break in the reading flow here we first talk about the nn libraries in jax and then proceed with describing a framework that does not use any of them!
While not strictly necessary to create a SNN library as neatly demonstrated by \emph{Rockpool}, they can massively streamline implementation and improve user experience.\\
\textit{Rockpool} \cite{muir2019_rockpool} is one of the earliest effort to leverage JAX for efficient SNN simulation.
Currently, Rockpool is used as a training platform for the accompanying Xylo SNN inference platform, and so its published functionality is limited to linear layers and LIF neuron dynamics.
%More elaborate building blocks such as convolutional layers were not available at the time of writing.
Likewise, \textit{JAXSNN} \cite{mueller2024jaxsnn} falls into this line of hardware-centric SNN libraries. 
JAXSNN is designed with a time-continuous approach to SNNs in mind, exploiting event-prop \cite{Wunderlich_Pehle21_evenback} and time-to-first-spike methods \cite{goltz2020fast} for gradient computation which tie seamlessly with the accompanying BrainScaleS-2 platform to enable hardware-in-the-loop training.
For this reason, it is not a general-purpose training framework that supports arbitrary connectivity and neuron types but rather focuses on maximally enhancing performance on the underlying platform.\\
Haiku and Flax \cite{haiku2020_github, flax2020_github} are frameworks that reimplement several elements of the JAX API such as batching, internal state management and random number generation.
\begin{figure}
\centering
  \includegraphics[width=0.43\textwidth]{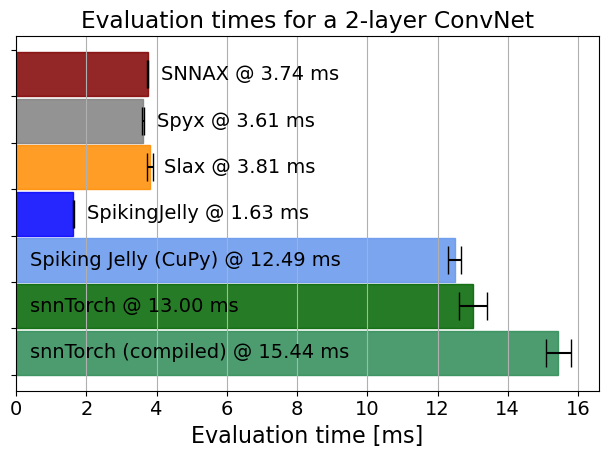}
\caption{Execution time measurements of a two-layer LIF-CNN with kernel size 3 on a 48x48 pixel image and batchsize 32.}
\label{fig:conv}
\end{figure}
One of the most recently published SNN libraries is Spyx \cite{heckel2024_spyx} builds on the Haiku library.
It is designed for maximum training speed on feed-forward architectures and achieves this through layer-by-layer execution and Haiku's loop unrolling feature.
This reduces the CPU-GPU communication overhead significantly and allows the batch-parallel application of stateless operations.
The downside of this method is the lack of support for arbitrary recurrent connectivity, which can be beneficial on some tasks.
Thanks to Haiku, Spyx supports a variety of neural network blocks and neuron models while still achieving similar performance to other frameworks that require the implementation of custom CUDA kernels.
It also supports the mapping of the trained parameters to neuromorphic hardware via the Neuromorphic Intermediate Representation (NIR) \cite{pedersen2023_nir}.
In a similar vein, Slax \cite{summe2024slax} builds on Flax and is geared towards online training and algorithmic exploration.
With this goal in mind, Slax not only supports BPTT but various other approximations like OSTL, RTRL, OTTT, OTPE and FPTT.
As opposed to Spyx, Slax supports almost arbitrary connectivity thereby enabling recurrent connections.
This comes at the price of step-by-step execution of the entire network architecture as opposed to Spyx' layer-by-layer approach.
%This can also limit the support for nesting of JAX functions and neural network layers.
Flax and Haiku come with the downside that users interested in exploring beyond provided models are required to learn Haiku or Flax, and a much less intuitive interface compared to PyTorch.
In contrast, Equinox is a minimal library that offers many scientific computation and neural network features, while being fully compatible with all existing JAX functionalities. 
SNNAX builds on Equinox as underlying deep learning library.
Like Spyx, this allows building state-of-the-art networks with custom building blocks in a way similar to PyTorch modules while  operating seamlessly with JAX functions.
We believe that Equinox' approach is more suitable for the cutting-edge algorithmically-minded user that wishes to explore new brain-inspired algorithms and structures.
Many of the discussed libraries share a unified interface provided by the NIR initiative which enables a certain level of cross-compatibility between them thus also partly exposing the interfaces of a number of neuromorphic accelerators.
\section{A Machine Learning Library for \acp{SNN}}
While developing SNNAX, we identified a set of key functionalities that a SNN library should satisfy to efficiently train modern spiking neural networks.
In the following sections, we will discuss these key functionalities and describe their implementation in SNNAX.

\subsection{Network Connectivity and Hardware Acceleration}
A key element in modern deep learning is the specification of the connectivity between neurons.
This is typically realized by grouping neurons into layers whose connectivity is then described by highly regular connections, e.g. dense all-to-all connections or convolutions.
SNNAX is built on Equinox, a JAX library that already provides the core functionalities to describe the network connectivity. 
Since the corresponding mathematical operations are very costly, but also easily parallizable, SNNAX exploits JAX' XLA backend to run these on dedicated accelerators such as GPUs and TPUs.
XLA (aXelerated Linear Algebra) \cite{leary2018xla} is an increasingly popular ML compiler that was originally designed as Tensorflow backend, but now also serves as the backbone of JAX. 
It defines a representation of the computational graph, optimizes it, and deploys it onto the target hardware.
Furthermore, XLA is particularly amenable to extension with novel devices and computing architectures as demonstrated in \cite{Lohoff_etal23_inteneur}.
% Building on XLA, JAX provides a reliable JIT compilation feature that vastly accelerates common deep learning workloads.
%%Emre: explain here to the layman what jit compilation means, why does it accelerate?
SNNAX and Equinox are both able to fully harness JAX' JIT compilation, making it a competitive choice for efficient SNN evaluation.
In fact, both support all of JAX' function transformation including automated and intuitive batching through \textbf{vmap} and device parallelism through \textbf{pmap} and array sharding mechanism.
% Additionally SNNAX supports JAX' experimental data parallel framework (sharding), enabling near seamless distribution across GPUs on a node.

\subsection{Stateful Layers and Long Sequences}
SNNs are temporally evolving dynamical systems modeled through differential equations.
Discretized \acp{SNN} can be viewed as a special type of \ac{RNN} with multiple internal states, akin to an \ac{LSTM} \cite{Zenke_Neftci21_brailear}. 
In its most general case of relevance to the implementation in ML frameworks, \ac{SNN} dynamics can be written as:
\begin{equation}\label{eq:snn}
  \begin{aligned}
    U^{t+\Delta t}_i &= f(U^t_i, S^t_i, \theta) + g(U^t, S^t, \phi); \, t=0,...,T\Delta t \\
    S^t_i &= \Theta(U^t_i)
  \end{aligned}
\end{equation}
where \(U^t \in \mathbb{R}^{N\times M}\) is the internal state of the neuron which consists of \(M\) states (\emph{e.g.} compartments) and \(S^t \in \mathbb{R}^{N}\) is the output of other neurons activity (including inputs). 
Additionally, \(\Theta\) is a discontinuous threshold function, \(\theta\) and \(\phi\) are learnable parameters of the neuron dynamics, and \(T\) is the number of time steps. \(N\) denotes the number of neurons and \(M\) denotes the number of states within each neuron.
Because the dynamics above often result from the discretization of continuous dynamics, \(\Delta t\) is generally small, leading to \(T\) being large (generally \(T > 100\)) compared to \acp{RNN} (generally \(N < 10\)).
Dealing with the internal states of the neurons is often tedious within JAX' functional programming paradigm.
SNNAX therefore handles the state management behind a thin layer that is still easily accessible to the user.
% In our opinion, this is a key feature for the algorithmically inclined researcher as it enables more fine-grained control over the evaluation and training of the network, and the monitoring of intermediate states.

\subsection{Gradient-based Learning}
Although Neuroscience provides principled models of biological learning, the state-of-the-art for training SNN for most practical applications is still gradient descent.
There, automatic differentiation (AD) and the backpropagation algorithm are used to compute the gradient of a cost function with respect to the models parameters \cite{Baydin_etal17_autodiff}.
AD allows computation of gradients up to machine precision by decomposing the models computational graph into its elemental operations whose symbolic derivatives are known and then utilizes the chain rule to assemble the functions that computes the derivatives.
% JAX provides cutting-edge AD features that can be leveraged when tackling SNN which are known to be notoriously difficult to train, such as gradient checkpointing and forward-mode AD.

\subsubsection{Temporal Credit Assignment}
SNNAX maps spikes into a regular temporal grid through spike time binning, which makes it functionally similar to snnTorch and Spyx.
It then treats the discretized form of the SNNs ODE as an RNN which is evaluated using a \textbf{for}-loop over the time dimension with changing input and internal state but fixed computational graph per time step.
The correct implementation of the temporal processing becomes particularly relevant when considering layer-by-layer and step-by-step execution of spiking neural network architectures.
Unrolling the \emph{jax.lax.scan} primitive for several steps instead of just one can be beneficial in the layer-by-layer case because it reduces the CPU-GPU communication and allows the batch-parallel application of stateless operations such as matrix-multiplications and convolutions which can incur massive runtime improvements for feed-forward architectures.
SNNAX, evaluates the time-loop over the entire network in a step-by-step manner as opposed to the layer-by-layer approach in Rockpool or Spyx.
While this is a performance bottleneck in some cases, we believe that this design choice is necessary since it enables recurrent connections across layers, a key feature of bio-plausibe SNNs that is paramount for the successful implementation of learning algorithms such as e-prop \cite{Bellec_etal20_soluto}.\\
% Since Python's \textbf{for}-loops are slow due to the global interpreter lock and the replication of the entire model along its temporal axis (loop unrolling), 
% JAX provides optimized loop primitives (\emph{i.e.} scan) that allows more efficient evaluation of for-loops that avoid loop unrolling.
Figures \ref{fig:mlp} and \ref{fig:conv} show execution times of LIF-based feed-forward MLP and CNN architectures for different SNN libaries on a single RTX 4090 GPU with a batchsize of 32.
%% other loop functions don't, k
% with the downside of requiring a fixed number of timesteps.
When differentiated, we arrive at a very efficient version of backpropagation through time (BPTT) \cite{Williams_Zipser95_gradlear}, the de-facto standard for training SNNs.

\subsubsection{Gradient Across Spikes}
A spiking neuron emits a spike when its internal state reaches a certain threshold. 
Once the threshold is crossed, the state generally resets to a new value, which makes spike emission a apparent non-differentiable process. 
Several approaches exist to circumvent the non-differentiable threshold with surrogate gradient approaches currently being the most popular solution \cite{Neftci_etal19_surrgrad}. 
There, the ill-defined derivative of the spiking function is replaced with a surrogate function that then allows to obtain a smooth gradient for the entire network.
Since SNNAX is built on JAX' advanced AD features, it is straightforward to define these surrogates and customize them according to the users requirements and several surrogate functions commonly found in literature are already implemented.
\begin{lstlisting}[caption={Implementation of a feed-forward SNN in SNNAX with tools from equinox and JAX. Note the consequent use of PyTree filters throughout the entire implementation.}, captionpos=b, label=lst:codeexample]
# ...
import equinox as eqx
import snnax.snn as snn

model = snn.Sequential(eqx.Conv2D(2, 32, 7, 2, key=key1),
                            snn.LIF((8, 8), [.9, .8], key=key2),
                            snn.flatten(),
                            eqx.Linear(64, 11, key=key3),
                            snn.LIF(11, [.9, .8], key=key4))
# ...
# Simple batched loss function
@partial(jax.vmap, in_axes=(0, 0, 0))
def loss_fn(in_states, in_spikes, tgt_class):
    out_state, out_spikes = model(in_states, in_spikes)

    # Spikes from the last layer are summed
    pred = out_spikes[-1].sum(-1)
    loss = optax.softmax_cross_entropy(pred, tgt_class)
    return loss

# Calculating the gradient with equinox PyTree filters and
# subsequently jitting the resulting function
@eqx.filter_jit
@eqx.filter_value_and_grad
def loss_and_grad(in_states, in_spikes, tgt_class):
    return jnp.mean(loss_fn(in_states, in_spikes, tgt_class))
# ...
# Simple training loop
for spikes, tgt_cls in tqdm(dataloader):
    # Initializing the membrane potentials of LIF neurons
    states = model.init_states(key)
    
    # Jitting with equinox PyTree filters
    loss, grads = loss_and_grad(states, spikes, tgt_cls)

    # Update parameter PyTree with equinox and optax
    updates, opt_state = optim.update(grads, opt_state)
    model = eqx.apply_updates(model, updates)
    
\end{lstlisting}
\subsubsection{Advanced AD Tools}
JAX' cutting-edge AD features enable a host of involved learning techniques that have proven to be helpful assets when training SNNs.
SNN training can be very memory-intensive due to the small simulation time steps necessary, often resulting in sequence lengths in the hundreds or even thousands.
When training with BPTT, this quickly saturates the memory of the underlying accelerator, and thus training large models can become prohibitively slow.
JAX, like PyTorch provides a gradient checkpointing tool that reduces memory consumption through recomputation during the backward pass.
Furthermore, JAX provides a fully-fledged forward-mode AD implementation that allows to trade memory for compute which enables the design of elaborate gradient-based learning algorithms that save on both memory and compute or might even enable online learning \cite{Zenke_Neftci21_brailear}.
% A dedicated module implements checkpointed versions of the scan loop, allowing to trade off runtime vs memory.
Furthermore, JAX allows, with only very few limitations, to compute arbitrarily high orders of derivatives, thereby facilitating Meta-learning and learning-to-learn approaches which have been demonstrated to be particularly useful for few-shot learning \cite{Stewart_Neftci22_metaspik}.
The implementation in SNNAX is straight forward in this case, by just differentiating the already differentiated code once again.

\subsection{User Interface}
SNNAX, much like Equinox, has been designed with user experience in mind.
A common motive of many JAX libraries is the extensive use of PyTrees, which are simple data structures that package large assemblages of arrays and related structures.
Equinox represents parameterised functions as immutable PyTree instances in order to be compatible with JAX' functional programming paradigm.
This also allows to directly use JAX' functional transformations like \emph{jax.jit} or \emph{jax.grad} on calls to the model's class instance without the need for further modifications as required by Haiku or Flax.
Both frameworks require additional steps like Flax' \emph{@nn.compact} decorator or Haiku's set of \emph{hk.transform} primitives which convert the object-oriented model implementation into an object that is compatible with JAX' functional programming paradigm.
These additional levels of abstraction can introduce unforseeable side-effects, incur additional implementation overhead or severely limit the functionality of certain features such as the use of JAX transformations within \emph{hk.transform}.
SNNAX embraces Equinox' \emph{Everything-is-a-PyTree} paradigm and provides a set of additional tree primitives and management tools dedicated to the management of stateful neural network models.
In this vein, SNNAX also provides a set of convenience function such as \textit{snnax.Sequential} to enable the rapid prototyping and layer-by-layer execution of feed-forward networks similar to Spyx (see Listing \ref{lst:codeexample}), while \textit{snnax.SequentialRecurrent} allows the implementation of recurrent models as demonstrated in Slax.
For more elaborate models with complex recurrent connections, SNNAX provides a \textit{GraphStructure} class that allows the definition of intricate feedback loops and recurrences through a graph object.
Alternatively, users can implement custom behavior by implementing their own stateful layers based on the \textit{StatefulLayer} class.
The graph representation is then meticulously executed by SNNAX' execution loop while being fully differentiable.
Finally, since Equinox attempts to mimic the simplicity of the PyTorch API, SNNAX inherits many of these features as well which flattens the learning curve and reduces the codebase footprint significantly.

\section{Conclusion}
We introduced SNNAX as a new, fast and user-friendly SNN simulation library that allows to efficiently train modern SNNs.
Its design principles are in line with already existing SNN libraries and it matches their performance without the need for writing custom CUDA code while being easy to maintain and extend.
% For the future, we plan to include an option that allows SNNAX to use layer-by-layer execution when the underlying network architecture permits it.
% For this purpose we will exploit the representation of the architecture as a graph object.
% Furthermore, we intend to leverage advanced AD methods to implement a fully online version of e-prop \cite{Bellec2020eprop} and intend to add an additional module to SNNAX that allows to efficiently train spike-time based learning algorithms such as event-prop based on Equinox' accompanying library for numerical differential equation solving called \emph{diffrax} \cite{kidger2021on}.
% Finally, we hope to integrate our work into the NIR framework which would enable it to be part of the neuromorphic community and its vast hardware and software ecosystem.
% Once integrated, we believe that SNNAX is poised to tackle the coming challenges of neuromorphic computing.

\section*{Acknowledgements}
This work was sponsored by the Federal Ministry of Education, Germany (project NEUROTEC-II grant no. 16ME0398K and 16ME0399) and NeuroSys as part of the initiative "Clusters4Future", funded by the Federal Ministry of Education and Research BMBF (03ZU1106CB).
We also would like to thank Anil Kaya and Anurag K. Mishra for their valuable contributions to the development of the framework.

\bibliographystyle{IEEEtran}
\bibliography{refs, biblio_unique_alt}

\end{document}

%% file: acronyms.tex
\acrodef{IR}[IR]{Intrinsic Rewards and Motivation}
\acrodef{PPO}[PPO]{Proximal Policy Optimization}
\acrodef{RL}[RL]{Reinforcement Learning}
\acrodef{AC}[AC]{Arrenhius \& Current}
\acrodef{AD}[AD]{Automatic Differentiation}
\acrodef{AER}[AER]{Address Event Representation}
\acrodef{AEX}[AEX]{AER EXtension board}
\acrodef{AMDA}[AMDA]{``AER Motherboard with D/A converters''}
\acrodef{ANN}[ANN]{Artificial Neural Network}
\acrodef{API}[API]{Application Programming Interface}
\acrodef{BP}[BP]{Back-Propagation}
\acrodef{BPTT}[BPTT]{Back-Propagation-Through-Time}
\acrodef{BM}[BM]{Boltzmann Machine}
\acrodef{CAVIAR}[CAVIAR]{Convolution AER Vision Architecture for Real-Time}
\acrodef{CCN}[CCN]{Cooperative and Competitive Network}
\acrodef{CD}[CD]{Contrastive Divergence}
\acrodef{CG}[CG]{Computational Graph}
\acrodef{CMOS}[CMOS]{Complementary Metal--Oxide--Semiconductor}
\acrodef{CNN}[CNN]{Convolutional Neural Network}
\acrodef{COTS}[COTS]{Commercial Off-The-Shelf}
\acrodef{CPU}[CPU]{Central Processing Unit}
\acrodef{CV}[CV]{Coefficient of Variation}
\acrodef{CTC}[CTC]{connectionist temporal classification}
\acrodef{DAC}[DAC]{Digital--to--Analog}
\acrodef{DBN}[DBN]{Deep Belief Network}
\acrodef{DCLL}[DECOLLE]{Deep Continuous Local Learning}
\acrodef{DFA}[DFA]{Deterministic Finite Automaton}
\acrodef{DFA}[DFA]{Deterministic Finite Automaton}
\acrodef{divmod3}[DIVMOD3]{divisibility of a number by 3}
\acrodef{DPE}[DPE]{Dynamic Parameter Estimation}
\acrodef{DPI}[DPI]{Differential-Pair Integrator}
\acrodef{DSP}[DSP]{Digital Signal Processor}
\acrodef{DVS}[DVS]{Dynamic Vision Sensor}
\acrodef{EDVAC}[EDVAC]{Electronic Discrete Variable Automatic Computer}
\acrodef{EIF}[EI\&F]{Exponential Integrate \& Fire}
\acrodef{EIN}[EIN]{Excitatory--Inhibitory Network}
\acrodef{EPSC}[EPSC]{Excitatory Post-Synaptic Current}
\acrodef{EPSP}[EPSP]{Excitatory Post--Synaptic Potential}
\acrodef{eRBP}[eRBP]{Event-Driven Random Back-Propagation}
\acrodef{FPGA}[FPGA]{Field Programmable Gate Array}
\acrodef{FSM}[FSM]{Finite State Machine}
\acrodef{FZJ}[FZJ]{Foschungszentrum J\"ulich}
\acrodef{GPU}[GPU]{Graphical Processing Unit}
\acrodef{HAL}[HAL]{Hardware Abstraction Layer}
\acrodef{HH}[H\&H]{Hodgkin \& Huxley}
\acrodef{HMM}[HMM]{Hidden Markov Model}
\acrodef{HNF}[HNF]{Helmholtz Nano-Facility}
\acrodef{HW}[HW]{Hardware}
\acrodef{hWTA}[hWTA]{Hard Winner--Take--All}
\acrodef{ID}[ID]{Implicit Differentiation}
\acrodef{IF2DWTA}[IF2DWTA]{Integrate \& Fire 2--Dimensional WTA}
\acrodef{IF}[I\&F]{Integrate \& Fire}
\acrodef{IFSLWTA}[IFSLWTA]{Integrate \& Fire Stop Learning WTA}
\acrodef{INCF}[INCF]{International Neuroinformatics Coordinating Facility}
\acrodef{INRC}[INRC]{Intel Neuromorphic Research Community}
\acrodef{INM}[INM]{Institute of Neuroscience and Medicine}
\acrodef{INI}[INI]{Institute of Neuroinformatics}
\acrodef{IO}[IO]{Input-Output}
\acrodef{IoT}[IoT]{internet of things}
\acrodef{IPSC}[IPSC]{Inhibitory Post-Synaptic Current}
\acrodef{IPU}[IPU]{Intelligence Processing Unit}
\acrodef{ISI}[ISI]{Inter--Spike Interval}
\acrodef{JFLAP}[JFLAP]{Java - Formal Languages and Automata Package}
\acrodef{JIT}[JIT]{just-in-time compilation}
\acrodef{JSC}[JSC]{J\"ulich Supercomputing Center}
\acrodef{LIF}[LIF]{Linear Integrate and Fire}
\acrodef{LSM}[LSM]{Liquid State Machine}
\acrodef{LTD}[LTD]{Long-Term Depression}
\acrodef{LTI}[LTI]{Linear Time-Invariant}
\acrodef{LTP}[LTP]{Long-Term Potentiation}
\acrodef{LTU}[LTU]{Linear Threshold Unit}
\acrodef{LSTM}[LSTM]{long short-term memory}
\acrodef{MAML}[MAML]{Model Agnostic Meta Learning}
\acrodef{MCMC}{Markov Chain Monte Carlo}
\acrodef{MSE}{Mean-Squared Error}
\acrodef{NAS}[NAS]{Neural Architecture Search}
\acrodef{NHML}[NHML]{Neuromorphic Hardware Mark-up Language}
\acrodef{NMDA}[NMDA]{NMDA}
\acrodef{NME}[NE]{Neuromorphic Engineering}
\acrodef{PC}[PC]{Predictive Coding}
\acrodef{PCB}[PCB]{Printed Circuit Board}
\acrodef{PGI}[PGI]{Peter Gr\"unberg Institute}
\acrodef{PRC}[PRC]{Phase Response Curve}
\acrodef{PSC}[PSC]{Post-Synaptic Current}
\acrodef{PSP}[PSP]{Post--Synaptic Potential}
\acrodef{RI}[KL]{Kullback-Leibler}
\acrodef{RRAM}[RRAM]{Resistive Random-Access Memory}
\acrodef{RBM}[RBM]{Restricted Boltzmann Machine}
\acrodef{RTRL}[RTRL]{Real-Time Recurrent Learning}
\acrodef{ROC}[ROC]{Receiver Operator Characteristic}
\acrodef{RSA}[RSA]{Representational Similarity Analysis}
\acrodef{RDA}[RDA]{Representational Dissimilarity Analysis}
\acrodef{RDM}[RDA]{Representational Dissimilarity Matrix}
\acrodef{RNN}[RNN]{Recurrent Neural Network}
\acrodef{SAC}[SAC]{Selective Attention Chip}
\acrodef{SCD}[SCD]{Spike-Based Contrastive Divergence}
\acrodef{SCX}[SCX]{Silicon CorteX}
\acrodef{SG}[SG]{Surrogate Gradient}
\acrodef{SGD}[SGD]{Surrogate Gradient Descent}
\acrodef{SRM}[SRM]{Spike Response Model}
\acrodef{SNN}[SNN]{Spiking Neural Network}
\acrodef{STDP}[STDP]{Spike Time Dependent Plasticity}
\acrodef{SW}[SW]{Software}
\acrodef{sWTA}[SWTA]{Soft Winner--Take--All}
\acrodef{TPU}[TPU]{Tensor Processing Unit}
\acrodef{VHDL}[VHDL]{VHSIC Hardware Description Language}
\acrodef{VLSI}[VLSI]{Very  Large  Scale  Integration}
\acrodef{WTA}[WTA]{Winner--Take--All}
\acrodef{XML}[XML]{eXtensible Mark-up Language}
\acrodef{SIMD}[SIMD]{Single Instruction Multiple Data}
\acrodef{MIMD}[MIMD]{Multiple Instruction Multiple Data}
\acrodef{UCI}[UCI]{University of California Irvine}
\acrodef{Op}[Op]{Operation}
\acrodef{ISA}[ISA]{Instruction Set Architecture}
\acrodef{MLP}[MLP]{Multilayer Perceptrons}
\acrodef{XLA}[XLA]{Accelerated Linear Algebra}
\acrodef{ML}[ML]{machine learning}
\acrodef{MLF}[MLF]{machine learning framework}